# Overcoming Complexity Catastrophe: An Algorithm for Beneficial Far-Reaching Adaptation under High Complexity

*Sasanka Sekhar Chanda[1], Sai Yayavaram[2]

**Affiliations:**

[1]Department of Strategic Management, Indian Institute of Management Indore.

[2] Strategy Department, Indian Institute of Management Bangalore.

*Correspondence to: sschanda@iimidr.ac.in[1] sai.yayavaram@iimb.ac.in[2]

**Abstract:** In his seminal work with *NK* algorithms, Kauffman noted that fitness outcomes from algorithms navigating an *NK* landscape show a sharp decline at high complexity arising from pervasive interdependence among problem dimensions. This phenomenon—where complexity effects dominate (Darwinian) adaptation efforts—is called *complexity catastrophe*. We present an algorithm—incremental change taking turns (*ICTT*)—that finds distant configurations having fitness superior to that reported in extant research, under high complexity. Thus, complexity catastrophe is not inevitable: a series of incremental changes can lead to excellent outcomes.

___________________________________________________________________

Search for high-fitness peaks on Kauffman's *NK* model landscape has been invoked by researchers in multiple fields (*1*), for example physics (*2*), evolutionary biology (*3*), complexity theory (*4*) management (*5*), artificial life (*6*) etc. In the *NK* model the fitness of an *N*-node decision configuration is the average of fitness contributions across individual nodes; the fitness contribution[1] by an individual node depends on its own state value ("0" or "1") and on the state values ("0" or "1") in *K* other nodes on which the focal node has a dependency. The parameter *K* is a measure of complexity arising from interdependence. [2]

---

[1] Fitness contribution values are randomly drawn from the *Uniform Distribution* U (0, 1).
[2] We elaborate further on the *NK* model in the *Supplementary document*.

Typical algorithms in extant research (*2, 6-9*) report a precipitous drop in fitness attainable, as complexity increases. Kauffman (*3*, *4*) cites this phenomenon as *complexity catastrophe.* We present an algorithm—**incremental changes, taking turns** (*ICTT*)—where this drop is much lower. In other words, under high complexity arising from pervasive interdependence between problem dimensions, *ICTT* attains fitness superior to that reported by Kauffman (*2*, *3*), thereby contesting the inevitability of *complexity catastrophe*. This provides a powerful tool to scientists and decision-makers working on phenomena involving high complexity. Moreover, compared to other algorithms, the *ICTT* finds excellent peaks further from the starting position. This makes it an appropriate vehicle for bringing about far-reaching adaptation.

**Baseline algorithms for fitness walk on the *NK* landscape**

In order to highlight *complexity catastrophe*, Kauffman uses a **simple model of myopic local search** (*8*, *10*), which is also referred to as **centralized search** (**CS**) (*7*, *9*). At initialization, nodes of an *N*-bit decision string are randomly assigned state values of "0" and "1" with equal probability (0.50). One randomly-chosen node is flipped in a given time step. The move is confirmed if overall fitness is higher, and rejected otherwise. Search terminates when it is not possible to find an immediate neighbor with higher fitness or after the execution of a pre-specified number of time steps. The **centralized search** (**CS**) algorithm demonstrates *complexity catastrophe* since fitness outcomes decline rapidly for higher values of *K* (complexity arising from pervasive interdependence). This constitutes the baseline for comparison of our results. We further consider another algorithm from Kauffman's research, **parallel updating** (**PU**) (*2*) as a second / additional baseline.

Kauffman and colleagues (*2*) describe a *parallel updating* (**PU**) algorithm to find peaks superior to those obtainable from **CS**. Decision elements (nodes) are accorded a certain probability, $\tau$ ($0< \tau <1$) of flipping. In a given generation, all *N* decision elements attempt to flip in parallel, with probability τ. However, the nodes that are actually allowed to flip are the ones where overall higher fitness is accomplished if solely that node flipped. The process is continued for a pre-specified number of generations, or till a point is reached when it is not possible to obtain higher fitness by flipping one node. For a given combination of *N* and *K*, the value of τ for which **PU** performs the best is found by trial and error. We use τ = 0.33 and obtain fitness values close to the maximum obtainable across all values of *K*.

**Design of the ICTT Algorithm**

The *ICTT* algorithm is purposed with finding high fitness configurations for an *N*-bit decision configuration, on Kauffman's *NK* landscape. At initialization, we distribute the set of *N* decision elements into a number of sub-units roughly equally. If *N* is not divisible by the number of sub-units, first the maximum feasible number of nodes—chosen randomly—are equally distributed among the sub-units; thereafter remaining nodes are allocated to a randomly-chosen sub-unit.

In one variant of *ICTT*, **ICTT1**, we use four sub-units. In a second variant, **ICTT1_alt**, we use six sub-units. As in the previous cases, in order to derive the starting configuration, nodes of the *N*-bit decision string are randomly assigned state values of "0" and "1" with equal probability (0.50). In each time step a node is selected at random, for flipping. A move is made if the sum of fitness contributions of the co-members in the sub-unit where the focal node resides is higher than the corresponding value prior to the move; otherwise, a different node is flipped in the next time step, and the calculation is repeated. The highest fitness decision configuration encountered is the designated outcome of the search process. In effect, selection of configurations to move to is done by limited consideration of fitness contribution of a sub-unit. A decision configuration is *committed to* only when the fitness is higher than the fitness at the prior committed position. In the flow charts in **Figures 1** and **2** we present the logic above in a formal way.

**Figure 1**. Flowchart for the algorithm *incremental changes, taking turns (ICTT)*.

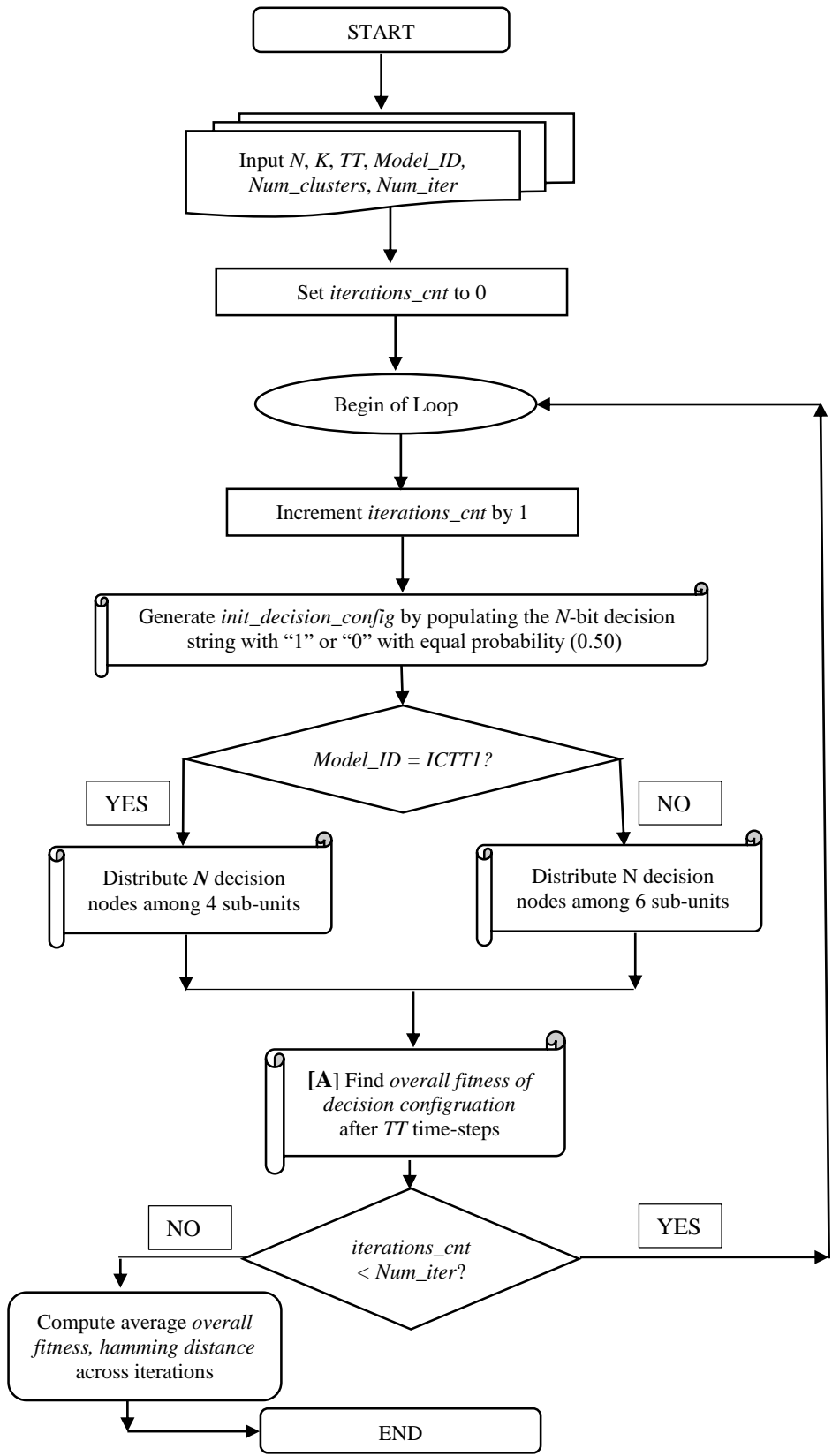

**Figure 2**. Flowchart of Block **A** of figure **1**: Processing in each time step in *ICTT*.

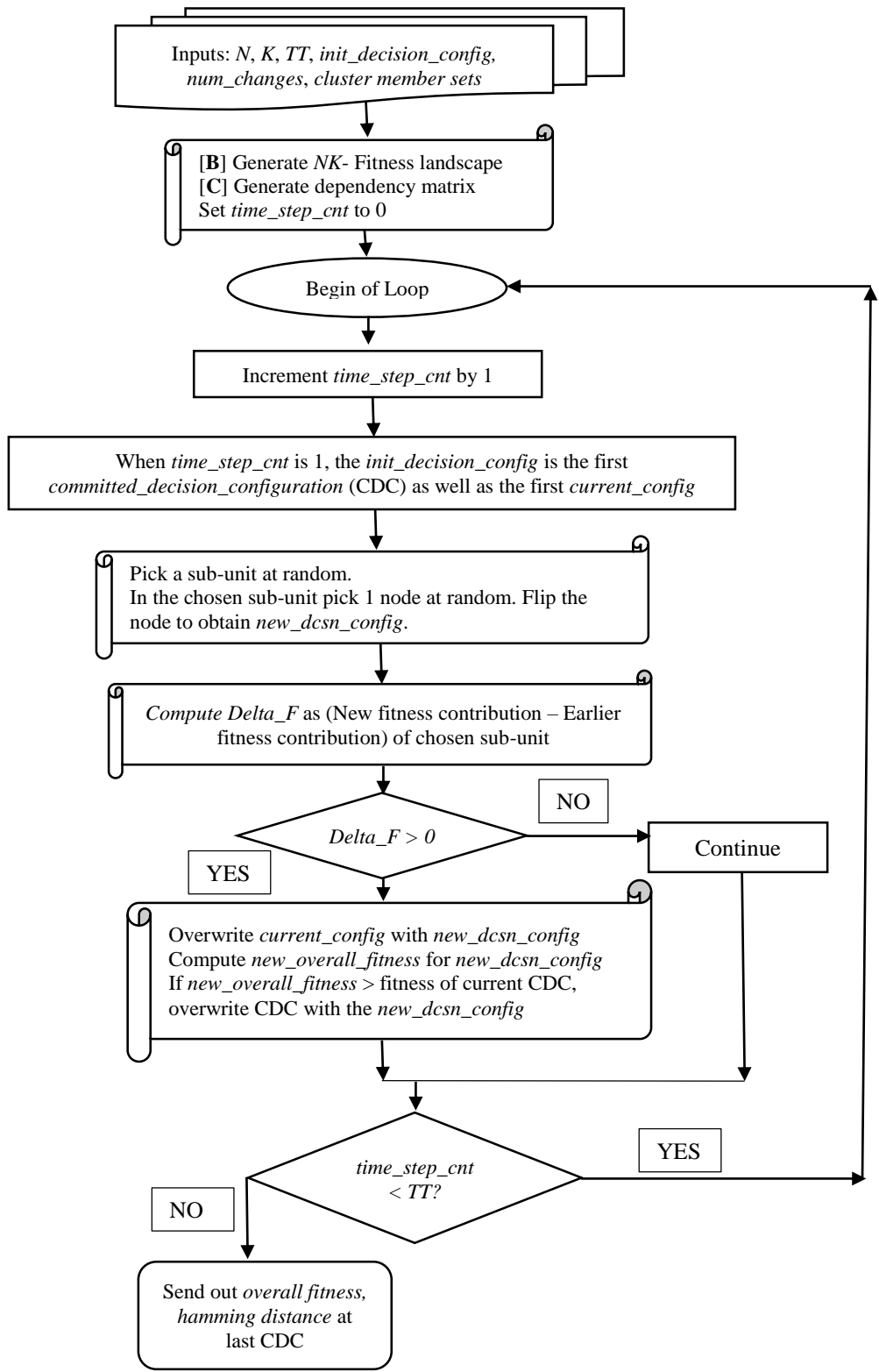

**[B]** A fitness landscape is a collection of $2^{(K+1)}$ rows and *N* columns of random numbers drawn from the *Uniform Distribution*, **U** (0, 1).

**[C]** In the dependency matrix each node is connected with *K* randomly chosen nodes. The fitness contribution of the focal node is a function of its own state value (0 or 1), and the state values (0 or 1) in those *K* other nodes.

## Results

In the results that follow we compare outcomes of *ICTT* variants (**ICTT1** and **ICTT1_alt**) with outcomes from two classical fitness walk algorithms **centralized search** (**CS**) (*7*, *9*) and **parallel updating** (**PU**) (*2*). All results shown are averages over experiments on 10,000 landscapes having $N = 20$ and run for $T = 1000$ time steps.

**Figure 3.** Comparison of **fitness** attained – *CS*, *PU* and *ICTT* variants

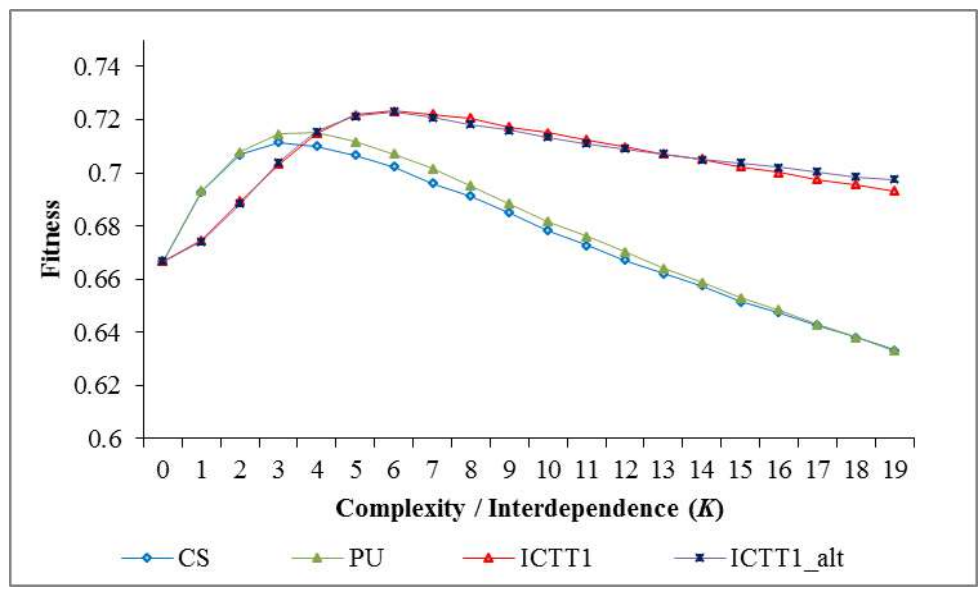

*Parameters.* $N = 20$; $T = 1000$; Iterations =10,000; **ICTT1** uses 4 sub-units; **ICTT1_alt** uses 6 sub-units; For **PU**, $\tau = 0.33$ and 500 generations are used.

In **Figure 3** we observe that while fitness outcomes from **CS** and **PU** algorithms fall off precipitously when complexity (or interdependence) $K$ is high, the decline is gentler for **ICTT1** and **ICTT1_alt**. Kauffman designates the steep (~8%) reduction in fitness between $K = 3$ and $K = 19$ for **CS** / **PU** as *complexity catastrophe* signifying failure of (Darwinian) adaptation in the face of rising complexity. For the *ICTT* variants, the corresponding fall in fitness is much lower (~1%). This suggests stymieing of *complexity catastrophe*.

In **Figure 4** we plot the extent of change from status quo, i.e., *hamming distance* between initial and final decision configurations. We observe that, compared to algorithms like **centralized search** and **parallel updating**, *ICTT* finds solutions that are more remote. In conjunction with the results from Figure **3** this suggests that *ICTT* is capable of bringing about beneficial radical change or far-reaching adaptation under mid and high complexity.

**Figure 4.** Comparison of **extent of change** accomplished – *CS*, *PU* and *ICTT* variants

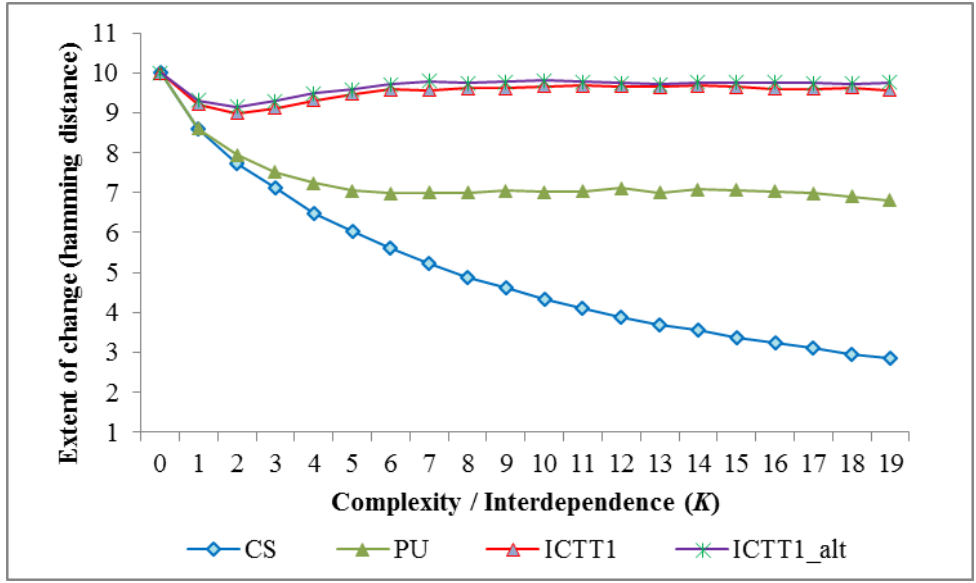

*Parameters.* ***N*** = 20; ***T*** = 1000; **Iterations** =10,000; **ICTT1** uses 4 sub-units; **ICTT1_alt** uses 6 sub-units; For **PU**, τ = 0.33 and 500 generations are used.

## Mechanism underlying success of *ICTT* at high complexity

In this section we explain why the **incremental changes, taking turns** (**ICTT1** variant) algorithm has sub-par outcomes under low complexity but excellent outcomes under high complexity. In Figure **5** we plot the *number of moves available* to **ICTT1** after each time step. This is computed as the number of nodes that can be flipped to increase fitness of the parent sub-unit. We observe that at low complexity (***K*** = 2) *ICTT* has progressively less moves for ***K*** = 2. Hence it quickly reaches a configuration where no further moves are feasible. This happens because the fitness landscape is highly correlated for low values of ***K*** (***K*** ≤ 3) (*11*). For high complexity (***K*** = 19) the crisis of running out of options happens much later or never happens at all. *ICTT* has many more moves available because the *NK* fitness landscape is practically uncorrelated for high ***K***. This enables *ICTT1* to survive longer and thereby find excellent peaks.

**Figure 5**. Comparison of **number of moves available** at low vs. high complexity, *ICTT1*

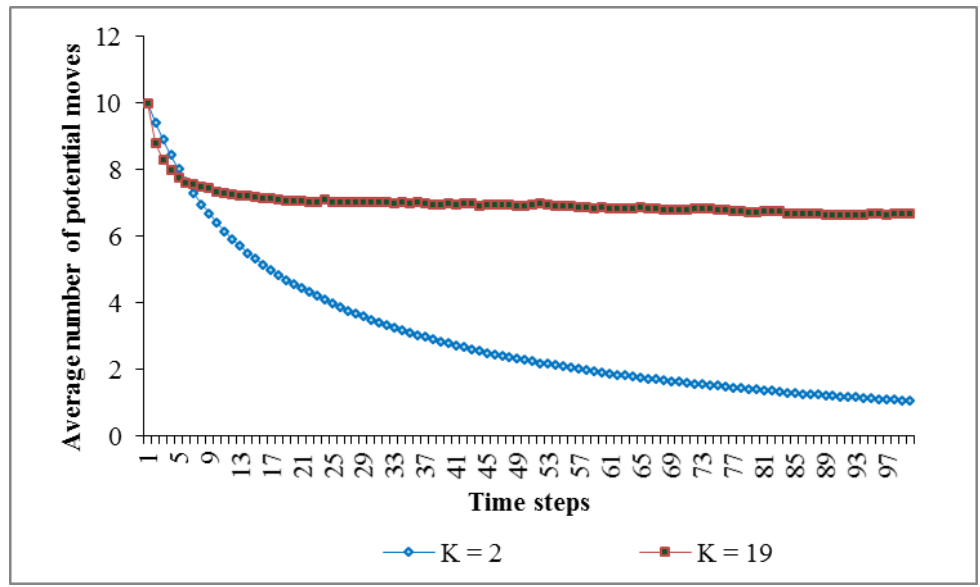

*Notes.* Rhombus-shaped markers represent search by **ICTT1** under conditions of low complexity (*K* = 2). Rectangular markers represent search by **ICTT1** under conditions of high complexity (*K* = 19).

*Parameters. N* = 20, **Time steps** = 100, **ICTT1** *is the ICTT variant utilizing four sub-units*.

## Discussion

The progress reported in this paper has applications in fields that invoke *NK* model for theoretical explanations. In general, we observe that beneficial radical change—or valuable far-reaching adaptation—is more likely to materialize by the approach of *ICTT* where (a) sub-units of an entity *take turns* to improve their contribution and (b) occasional drops in fitness during reconnaissance missions (venturing out from current committed position) is acceptable. Further, the superiority of *ICTT* under medium and high complexity (*K* > 5 in Figure 3) demonstrates that Kauffman's surmise—that "… breaking an organization into "patches" where each patch attempts to optimize for its own selfish benefit, even if that is harmful to the whole, can lead, as if by an invisible hand, to the welfare of the whole organization." (*4*, p. 247)—is sub-optimal at high complexity.

## Materials and Methods

<u>Kauffman's *NK* model</u>

In the *NK* model, a decision contains *N* decision elements or nodes. Each node can have a value of either "0" or "1". The *fitness contribution* by an individual node is jointly dependent on its value and the values in *K* other nodes with which the focal node shares a dependency. Higher the value of *K*, higher is the complexity arising from interdependence. A common problem formulated through the *NK* model concerns finding decision configurations with the highest (or lowest) **Fitness** values. This is considered an NP-hard problem because as *N* increases, the number of feasible decision configurations increases exponentially. For example, if $N = 16$, it is necessary to examine $2^N = 65,536$ decision configurations; for $N = 20$, it is necessary to examine over 1 million configurations.

A *decision configuration* is an instantiation of a decision, with all nodes filled with values of either "0" or "1". The extent of **Fitness** of a decision configuration is computed as the average of the *fitness contributions* of the *N* nodes making up a decision. To obtain the fitness contribution by a node, a matrix having $2^{(K+1)}$ rows and *N* columns—that we refer to as the *fitness matrix*—is populated with random draws from the *Uniform Distribution*. A numerical example will help clarify how an individual node's fitness is calculated.

Suppose that in a 5-bit decision-configuration ($N = 5$), the fitness contribution of any given node is dependent on its own value ("0" or "1") and the value in 3 other nodes ($K = 3$). Let us assume that we wish to know the fitness contribution of the 4th node ($p = 4$) by referring to the *fitness matrix* which is a table having $2^{(K+1)}$ rows and *N* columns filled with numbers randomly drawn from $U(0, 1)$, i.e. with random numbers between 0 and 1. We further assume that the fitness contribution 4th node depends on the values in the first, third and fifth nodes.

When the $p^{th}$ (4th) node has value 0, we refer to the upper $2^K$ rows of the fitness matrix. Else we refer to the lower $2^K$ rows of the fitness matrix. In the applicable section of the fitness matrix, we focus on the $p^{th}$ column, given that we wish to find the fitness contribution of the $p^{th}$ node. The row number where the relevant fitness value resides is [1 + the decimal equivalent of the binary number constructed by concatenating the state values of the *K* dependent nodes]. For example, if the state values of the first, third and fifth nodes are "0", "0" and "1" respectively, we look at the second row; if the state values are "0", "1" and "1" we look at the 4th row etc.